\documentclass[conference]{IEEEtran}
\IEEEoverridecommandlockouts
\PassOptionsToPackage{hyphens}{url}
\usepackage{cite}
\usepackage{dsfont}
\usepackage{amsmath,amssymb,amsfonts}
\usepackage{algorithmic}
\usepackage{graphicx}
\usepackage{textcomp}
\usepackage{xcolor}
\usepackage{tikz}
\usetikzlibrary{arrows.meta,positioning,fit,calc,shapes.multipart}
\usepackage{xcolor}
\usepackage{subcaption}   
\usepackage{graphicx}
\usepackage{dblfloatfix} 
\usepackage{comment}

\usepackage{ulem}

\usepackage{todonotes}

\newcommand{\modelname}{ViLiNT} 

\usepackage[colorlinks]{hyperref}

\begin{document}

\title{Multimodal embodiment\mbox{-}aware\\[-0.2ex]\mbox{navigation transformer}}

\author{
\IEEEauthorblockN{Louis Dezons\IEEEauthorrefmark{1}\IEEEauthorrefmark{2},
Quentin Picard\IEEEauthorrefmark{1},
Rémi Marsal\IEEEauthorrefmark{2},
François Goulette\IEEEauthorrefmark{2},
David Filliat\IEEEauthorrefmark{1}}
\IEEEauthorblockA{\IEEEauthorrefmark{1}AMIAD Pôle Recherche, Palaiseau, France}
\IEEEauthorblockA{\IEEEauthorrefmark{2}U2IS, ENSTA, Institut Polytechnique de Paris, Palaiseau, France}
\IEEEauthorblockA{\href{mailto:louis.dezons@polytechnique.edu}{louis.dezons@polytechnique.edu}}
}

\maketitle

\begin{abstract}

Goal-conditioned navigation models for ground robots trained using supervised learning show promising zero-shot transfer, but their collision-avoidance capability nevertheless degrades under distribution shift, i.e. environmental, robot or sensor configuration changes. We propose \modelname, a multimodal, attention-based policy for goal navigation, trained on heterogeneous data from multiple platforms and environments, which improves robustness with two key features. First, we fuse RGB images, 3D LiDAR point clouds, a goal embedding and a robot's embodiment descriptor with a transformer architecture to capture complementary geometry and appearance cues. The transformer's output is used to condition a diffusion model that generates navigable trajectories. 
Second, using automatically generated offline labels, we train a path clearance prediction head for scoring and ranking trajectories produced by the diffusion model. The diffusion conditioning as well as the trajectory ranking head depend on a robot's embodiment token that allows our model to generate and select trajectories with respect to the robot's dimensions.
Across three simulated environments, \modelname{} improves Success Rate on average by 166\%  over equivalent state-of-the-art vision-only baseline (NoMaD). This increase in performance is confirmed through real-world deployments of a rover navigating in obstacle fields.
These results highlight that combining multimodal fusion with our collision prediction mechanism leads to improved off-road navigation robustness\footnote{Code link: \url{https://github.com/LARIAD/ViLiNT}}.

\end{abstract}


\section{Introduction}

Robust autonomous navigation in off-road environments remains challenging due to the complexity and high variability of the environment, partial observation, dynamic obstacles, and sensing degradations (e.g., dust, rain, grass). Classical pipelines (e.g., \cite{roadrunner}) rely on  handcrafted or learned traversability maps and local planners. They offer interpretability and a good capacity to detect situations when motion is impossible. However, they may struggle when the deployment configuration (robot or sensor) changes, requiring at least sensor calibration or even new training data.  

End-to-end control trained by imitation (e.g., \cite{gnm}, \cite{vint}) has shown some robustness to the robot embodiment thanks to training on data from different robots and direct regression of controls or waypoints from onboard sensing. However, single-modality policies (typically RGB vision) may strongly degrade under environmental changes when other modalities like LiDAR may bring more robustness. These end-to-end local planners also often lack explicit mechanisms to deal with deadlocked situations, i.e., to exit local minima where no feasible trajectory to the goal is available. Finally, these approaches offer some robustness and adaptation to different embodiments, but they do not explicitly take the robot's dimensions into account and cannot easily be adapted to a new platform without fine tuning using data from this platform.

\begin{figure}[t]
  \centering
  \includegraphics[width=0.40\textwidth]{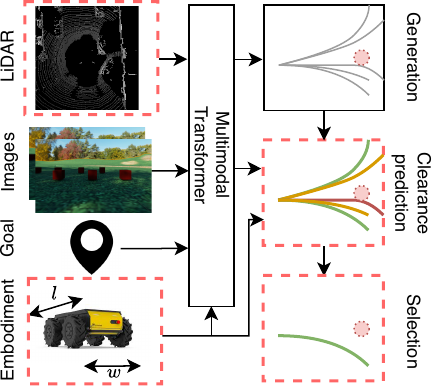}
  \caption{Overall architecture of our navigation model. Highlighted blocks in red indicate our main contributions. See text for details.}
  \label{fig:contribs}
\end{figure}

We therefore propose \textbf{\modelname}, a multimodal local motion planning architecture based on a transformer that fuses LiDAR, RGB images, 2D goal and embodiment data into a shared token space as shown in Fig. \ref{fig:contribs}.  This approach targets both \textit{navigation quality} by proposing efficient goal-reaching paths and \textit{safety} by predicting robot-specific safety margins.

Trajectories are generated with a goal-conditioned diffusion policy \cite{visuomotordiffusion} and fed to the clearance prediction head conditioned by the robot embodiment.  At inference, the system generates multiple candidate trajectories and selects the one with the minimum predicted risk (i.e. largest clearance), which is sent to the low-level controller. If no safe trajectory is found, an alternative sampling of exploratory trajectories is used to exit these local minima.

This design combines three complementary ideas: (i) multimodal token-level fusion to improve robustness; (ii) generative trajectory sampling to capture multi-modality in feasible plans; and (iii) an explicit clearance predictor trained with automatically generated labels from the training datasets.
Both the generator and the clearance predictor are parameterized by the robot size, the result is a motion-planner that is both expressive and selective: it can propose diverse options adapted to the robot embodiment and then reject those likely to collide, without requiring dense human labels. 

Our key contributions are: 
\begin{itemize}
  \item A \textbf{LiDAR--RGB tokenization and cross-modal Transformer} that integrates recent image frames, a LiDAR scan, goal coordinates, and platform description to produce robot's dynamics and scene description tokens.
  \item A \textbf{Navigation data augmentation} based on LiDAR data that enables clearance prediction learning without requiring data from all target platforms.
  \item A \textbf{Cross-attention clearance head} that uses per-waypoint queries (with robot-size conditioning) over LiDAR tokens to rank diffusion-generated trajectories to select the safest one.
 
\end{itemize}

\section{Related Work}
\label{sec:related}

\subsection{Perception and planning}
Many state-of-the-art traversable-path generation systems comprise two main components: (i) perception–mapping pipelines that estimate a traversability/cost field from onboard sensing, and (ii) model-based planners that optimize trajectories with respect to that field under kinematic/dynamic constraints. Observation and mapping pipelines infer traversability from RGB-D/LiDAR — often fusing proprioception for self-supervision — and recent works learn off-road costmaps and visual traversability that integrate with standard navigation stacks (e.g.,~\cite{traversability-costmap2}). On the planning side, reactive local planners such as 
TEB~\cite{teb}, as well as trajectory-optimization/Model Predictive Control (MPC) methods (e.g., MPPI~\cite{mppi}), remain strong baselines for smooth, constraint-aware motion. 
While such modular stacks are effective and interpretable, their performance typically hinges on \textit{careful platform-specific parameterization} (e.g., dynamic limits, and sensor-specific parameters). As sensing setups and robot dimensions change, these parameters often need substantial tuning, and robustness can degrade under appearance/geometry shifts despite good nominal performance. Motivated by these limitations, recent work has explored learning richer priors from data to \textit{reduce} manual cost design while keeping compatibility with classical controllers~\cite{vstrong}. Our approach follows this trend: we encode observations, robot embodiment, and goals with a Transformer and generate trajectories via a diffusion policy, while remaining compatible with a robot-specific low-level controller.

\subsection{End-to-end navigation models} 
Vision-only end-to-end driving systems such as DAVE-2~\cite{dave2} demonstrated that imitation from raw monocular imagery can produce lane-following and steering on-road, establishing the paradigm of mapping pixels to controls without intermediate costmaps. Building on this idea, goal-conditioned policies trained on broad, heterogeneous datasets (e.g., GNM/ViNT and follow-ups~\cite{gnm,vint,nomad}) show cross-robot/environment generalization and can output feasible trajectories, thereby reducing manual cost engineering while staying compatible with low-level controllers. However, purely visual policies degrade when observations lack salient features or contain unobserved regions due to limited field of view, which impacts collision-avoidance robustness under domain shift. LiDAR-centric approaches such as DTG~\cite{dtg} alleviate some visual brittleness but face sparsity and semantic ambiguity (e.g., thin obstacles). Both NoMaD~\cite{nomad} and DTG~\cite{dtg} employ diffusion models~\cite{visuomotordiffusion} to generate trajectories: DTG primarily uses diffusion as a distribution estimator, whereas NoMaD ranks candidate trajectories (e.g., by goal proximity) to select long-range plans—implicitly assuming generated trajectories are safe to track. We adopt diffusion for trajectory generation but pair it with an explicit clearance prediction head conditioned on multimodal tokens and embodiment, scoring candidates to improve robustness.

\subsection{Multimodal navigation models}

For \textit{multimodal sensing} in off-road navigation, Bird's-Eye-View (BEV) fusion networks trained with self-supervision predict traversability and geometry at low latency; e.g., RoadRunner~\cite{roadrunner} learns hindsight-labeled BEV traversability/elevation from camera and LiDAR. Vision-only baselines such as TerrainNet~\cite{terrainnet} produce semantic–geometric BEV maps that pair well with planners and are often combined with LiDAR in practice for robustness. Early multimodal end-to-end policies (e.g.,~\cite{agile}) fuse LiDAR and camera to learn direct control, demonstrating feasibility with low-cost sensors but exposing brittleness to distribution shift. Beyond BEV fusion, multimodal Transformers such as PACT~\cite{pact} use modality-specific encoders (e.g., PointNet~\cite{pointnet} for point clouds) to produce aligned tokens for sequence models that predict future states/actions. In line with these advances, we tokenize heterogeneous inputs and \textit{explicitly} encode robot size and goal in the token set; unlike direct-control approaches, we produce waypoint trajectories via diffusion, preserving compatibility across platforms.

\begin{figure*}[!t]
  \centering
  \includegraphics[width=\textwidth]{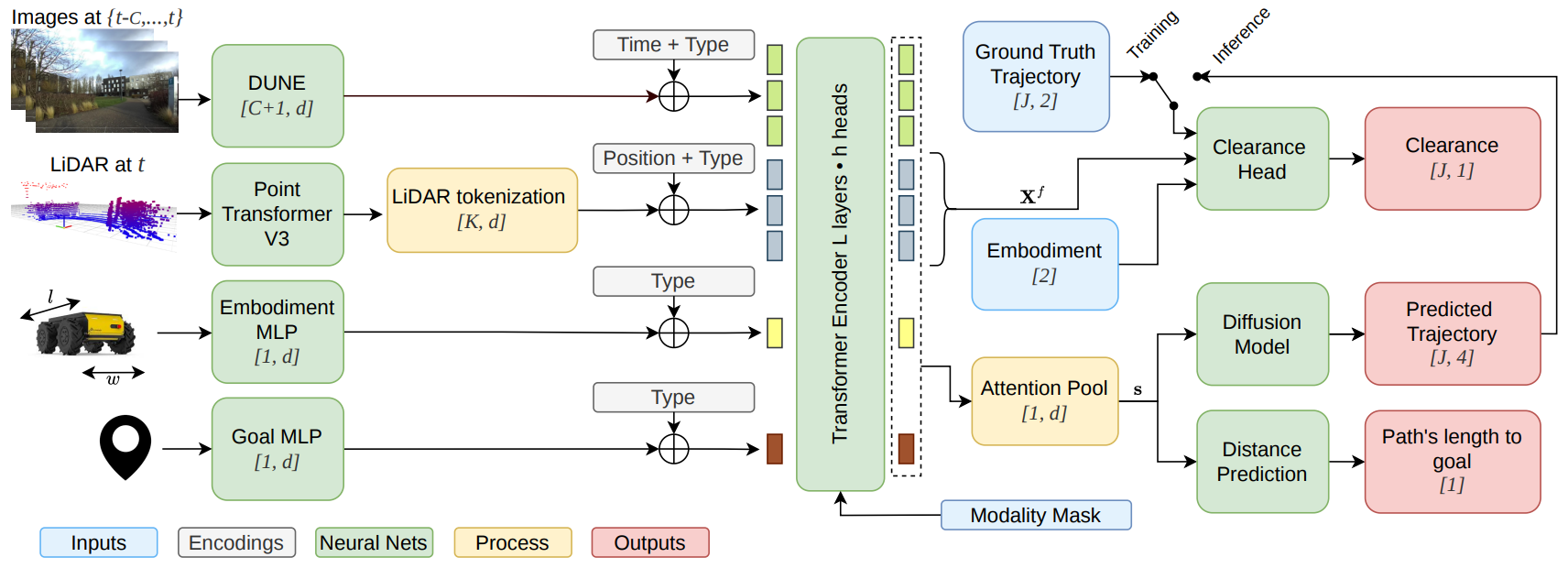}
  \caption{Global model architecture. See text for details.}
  \label{fig:architecture}
\end{figure*}

\subsection{Collision avoidance}
Several recent methods incorporate safety more explicitly. Risk-Guided Diffusion~\cite{risk_guided_diffusion} biases sampling with physics-based terrain risk to steer toward safer paths, at the cost of platform-specific modeling. RL-based diffusion planning~\cite{rl_diffusion_planning} integrates task constraints (goal reaching, obstacle avoidance) into diffusion training. MonoMPC~\cite{mono_mpc} learns an obstacle-clearance cost from monocular images and optimizes trajectories with MPC using that learned cost. Complementary to these directions, we learn a clearance head that scores diffusion proposals conditioned on LiDAR, embodiment, and goal, enabling risk-aware selection without hand-tuned costs.

\subsection{3D point cloud features encoding}
Point-cloud encoders for navigation and perception broadly fall into: \textit{(i) point-based} networks that operate on sets (e.g., PointNet/PointNet++), \textit{(ii) voxel-based} sparse 3D convolutions that scale to large scenes while preserving local geometry (e.g., Minkowski framework~\cite{minkowskiengine}), \textit{(iii) projection-based} encoders that trade some vertical detail for planar efficiency (e.g., polar/cylindrical discretizations~\cite{polarnet}) and \textit{(iV) point-transformers}.
Sparse 3D convs, such as MinkUNets~\cite{minkowskiengine}, provide a good trade-off between execution time and accuracy for downstream policy learning, however recent point-transformers~\cite{pointtransformerv3} offer increased accuracy, while being viable for embedded tasks with optimization such as Flash-Attention. 
We therefore use PointTransformerV3 as the 3D encoder with a polar sector–ring tokenization described in Sect.~\ref{subsec:lidar_tokenizer}.

\section{Method}
\label{sec:method}

Fig.~\ref{fig:architecture} illustrates our approach: sensor frames are encoded with DUNE \cite{dune} Vision Transformer for images, a state-of-the-art image encoder distilled from 2D and 3D visual features, and PointTransfomerV3 \cite{pointtransformerv3} for LiDAR data and then tokenized, augmented with type and temporal embeddings, and fused by a transformer that jointly attends to sensor tokens, a goal token and a robot's embodiment token. A conditional diffusion model samples multiple feasible waypoint sequences toward the goal. In parallel, a lightweight head conditioned on robot's size predicts the obstacle clearance around the trajectory for each candidate using automatically generated clearance values from the training datasets. Visual observation tokens are masked during training with a certain probability to enforce clearance prediction's reliance on 3D data. We predict the path's total length to the goal in a similar manner as Vint \cite{vint} predicts the time steps to go between the current robot's position and the goal's position to stabilize training. At inference, the system selects the trajectory with the maximum minimum distance to obstacles when some are detected nearby and the one bringing the robot the closest to the goal if not. Then, waypoints are sent to the low-level controller.

\subsection{LiDAR point cloud encoding and tokenization}
\label{subsec:lidar_tokenizer}

Let the instantaneous LiDAR point cloud be $\mathcal{P}=\{\mathbf{p}_i=(x_i,y_i,z_i)\}_{i=1}^N$. 
The set of points of size $N$ is fed to PointTransformerV3 \cite{pointtransformerv3} that returns per-point features $\mathbf{h}_i\in\mathbb{R}^{d_e}$, $i=1,..., N$. We denote $\mathcal{V}=\{(\mathbf{p}_i,\mathbf{h}_i)\}_{i=1}^N$ the resulting set (one feature per point). 

We partition the horizontal plane into a polar grid with $K_\theta$ angular sectors and $K_r$ radial rings over the region of interest (ROI) with horizontal field-of-view (FoV) $\theta\in[-\phi,\phi]$ and range $r\in[0,R]$, where $\phi$ is the half-FoV as described in Fig.~\ref{fig:lidar-tokenization}. 
We define sets $\mathcal{B}_{u,v}$ that contain each point of a given sector $(u,v)$ and the polar grid operator $\mathcal{B}$ that maps 2D points $(x_i, y_i)$ to their corresponding sector:
\begin{equation}
\begin{aligned}
\mathcal{B}_{u,v}
&= \big\{\, i \;:\; \mathcal{B}( x_i, y_i)=(u,v) \,\big\},\\
\mathcal{B}\;:&\; \mathbb{R}^{2}\to \{1,\ldots,K_\theta\}\times\{1,\ldots,K_r\}.
\end{aligned}
\end{equation}

\paragraph*{GeM pooling and empty-bin handling} The tokenizer outputs a fixed-budget set of $K=K_\theta K_r$ tokens by pooling encoder features within each cell. For a cell $(u,v)$ with features $\{\mathbf{h}_i\}_{i\in\mathcal{B}_{u,v}}$, we use Generalized-Mean (GeM) pooling, applied elementwise with learnable exponent $p>0$:
\begin{equation}
\mathbf{f}_{u,v}=\Big(\frac{1}{|\mathcal{B}_{u,v}|}\sum_{i\in\mathcal{B}_{u,v}}\mathbf{h}_i^{p}\Big)^{1/p} \in \mathbb{R}^{d_e},\label{eq:gem}
\end{equation}
If $\mathcal{B}_{u,v}=\varnothing$, we use a learned \textit{empty} embedding $\mathbf{e}_{\varnothing}\in\mathbb{R}^{d_e}$.

\begin{figure}[!t]
  \centering
  \includegraphics[width=0.4\textwidth]{figures/lidar_tokenization.png}
  \caption{Illustration of our LiDAR tokenization approach: PointTransformerV3 features are aggregated in a cylindrical voxel grid using Generalized Mean Pooling to produce tokens. Empty voxel tokens are represented by a learned feature vector.}
  \label{fig:lidar-tokenization}
\end{figure}

\paragraph*{Positional and range-aware descriptors} Each token is enriched with deterministic codes and simple statistics:
\begin{align}
\boldsymbol{\pi}_{\theta,u}&=[\sin\bar{\theta}_u,\cos\bar{\theta}_u],& \bar{\theta}_u&\text{: cell angle},\\
\boldsymbol{\pi}_{r,v}&=\bar r_v/R,& \bar r_v&\text{: cell radius},\\
\overline{r}_{u,v}&=\tfrac{1}{|\mathcal{B}_{u,v}|}\sum_{i\in\mathcal{B}_{u,v}} r(x_i, y_i),&  \overline{r}_{u,v}&\text{: mean range}, \\ 
\tilde r_{u,v}&=\sum_{i\in\mathcal{B}_{u,v}} w_i r(x_i, y_i),& \tilde r_{u,v}&\text{: soft-min center}
\end{align}

with near-range emphasis via normalized weights $w_i\propto e^{-\gamma r(x_i,y_i)}$. $\tilde r_{u,v}$ gives the token an explicit notion of "how close is the closest obstacle" in the sector which makes it relevant for collision detection.

\paragraph*{Final token projection} The token of dimension $d$ for cell $(u,v)$ is:
\begin{equation}
\begin{split}
\mathbf{x}_{u,v} &= W_f\,\mathbf{f}_{u,v} 
 \;+\; W_\pi\,[\boldsymbol{\pi}_{\theta,u};\boldsymbol{\pi}_{r,v}] \\
 &\quad +\; W_s\,[\overline{r}_{u,v};\tilde r_{u,v}]
 \;\in\mathbb{R}^{d},
\end{split}
\end{equation}
with learnable linear projections $W_f: \mathbb{R}^{d_e} \to \mathbb{R}^d,W_\pi: \mathbb{R}^3 \to \mathbb{R}^d,W_s: \mathbb{R}^2 \to \mathbb{R}^d$. We index sectors by sweeping counterclockwise around the robot’s forward axis. Rings are ordered from near to far.
This polar discretization balances non-uniform LiDAR densities and aligns with radial collision geometry, and relates to prior cylindrical encoders (e.g., \cite{polarnet}).

\subsection{Transformer scene encoding}
\label{subsec:scene_sa}

Given the current observation window, we form a token sequence by concatenating: (i) image-history tokens $\mathbf{I}\in\mathbb{R}^{(C+1)\times d}$ (one per frame from $t-C$ to $t$, obtained by the image encoder and linear projection), (ii) LiDAR tokens $\mathbf{X}\in\mathbb{R}^{K\times d}$ as described in Sect.~\ref{subsec:lidar_tokenizer}, (iii) a robot embodiment token $\mathbf{x}_{p}=E_\mathbf{w}(\mathbf{w})\in\mathbb{R}^d$, and (iv) a goal token $\mathbf{x}_{g}=E_g(\mathbf{g})\in\mathbb{R}^d$. Here, $E_\mathbf{w}: \mathbb{R}^{2} \to \mathbb{R}^d$ and $E_g: \mathbb{R}^{2} \to \mathbb{R}^d$ are single-layer projection MLPs, $\mathbf{g}$ is the goal position in the robot's frame and $\mathbf{w}=(w, \ell)$ the robot's embodiment vector with $w$ the robot's width and $\ell$ its length. The resulting sequence length is $S=(C+1)+K+2$.
We add (a) learned type embeddings to mark modality (image/LiDAR/size/goal), and (b) learned temporal index embeddings for the image tokens. Let $\mathbf{Y}\in\mathbb{R}^{S\times d}$ denote the sum of content, type, and positional encodings with the concatenated tokens. A Transformer encoder ($L$ layers, width $d$) computes fused representations:
\begin{align}
\mathbf{Z}^{0}&=\mathbf{Y},\\
\mathbf{Z}^{L}&=\operatorname{TF-Enc}\big(\mathbf{Y};\,\text{mask}=\mathbf{M}\big),
\end{align}
where $\mathbf{M}\in\{0,1\}^{S}$ is a per-token mask implementing goal masking and modality dropout during training. Masking allows the transformer not to rely on a single modality for conditioning the diffusion process and to predict clearance when generating the output tokens. We randomly mask the goal token in training to encourage feasible planning without the explicit goal, relying on scene tokens \cite{vint}, mitigating blocking failures induced by greedy pursuit of the goal.

\paragraph*{Scene token via attention pooling} 
We produce a single scene descriptor $\mathbf{s}\in\mathbb{R}^d$ by learned attention pooling over the final sequence $\mathbf{Z}^{L}=:(\mathbf{z}_1,\dots,\mathbf{z}_S)$:
\begin{equation}
\alpha_i=\operatorname{softmax}_i\Big(\tfrac{\mathbf{a}^\top \mathbf{z}_i}{\sqrt{d}}\Big),\qquad \mathbf{s}=\sum_{i=1}^S \alpha_i\,\mathbf{z}_i,\label{eq:attnpool}
\end{equation}
with a learned query vector $\mathbf{a}\in\mathbb{R}^d$. This pooling respects $\mathbf{M}$ by zeroing the scores of masked tokens. 

\paragraph*{Outputs for downstream heads} 
Besides $\mathbf{s}$, we expose the \textit{post-attention} LiDAR slice from $\mathbf{Z}^{L}$, $\mathbf{X}^f \in \mathbb{R}^{K \times d}$. The clearance prediction head can therefore query semantics-enriched fused LiDAR tokens while the trajectory generation diffusion head uses a pooled global scene and an image dynamics descriptor as conditioning. 

\subsection{Trajectory generation}
\label{subsec:waypoints}
We model a conditional distribution over a sequence of $J$ positional and angular waypoints (position relative to the robot and robot's $\cos$ and $\sin$ of the orientation difference) $\mathbf{u}_{1:J}\!\in\!\mathbb{R}^{J\times 4}$ given the fused context token $\mathbf{s}$  that summarizes observations, robot's size and 2-D goal direction. Following action-diffusion policies \cite{visuomotordiffusion}, we train a denoiser to invert a Gaussian noising process over the waypoint sequence. The timestep-conditioned ($t$) network $\varepsilon_\theta$ predicts the injected noise $\boldsymbol{\varepsilon}$ given the noisy sequence and the context $\mathbf{s}$:
\begin{align}
\hat{\boldsymbol{\varepsilon}} =\varepsilon_\theta\big(\mathbf{u}^t,\,t,\,\mathbf{s}\big),&\quad t=1,\dots ,T \\
\text{and we minimize}\quad 
\mathcal{L}_{\text{diff}} & =\mathbb{E}_{\mathbf{u}^0,\,t,\,\boldsymbol{\varepsilon}}\,\big\|\boldsymbol{\varepsilon}-\hat{\boldsymbol{\varepsilon}}\big\|_2^2.
\end{align}
The diffusion head outputs a distribution over waypoint sequences; at test time we sample $N$ candidates by running the reverse process for $T$ steps (we use $T=10)$.

For stabilizing training we also predict an auxiliary scalar for the distance to navigate until reaching the goal $d_g$ from $\mathbf{s}$ and apply an MSE loss:
\begin{equation}
    \mathcal{L}_{\text{aux}}=\big\|\hat{d}_g(\mathbf{s})-d_g\big\|_2^2
\end{equation}

\subsection{Clearance prediction}
\label{subsec:collision}

\begin{figure}[!t]
  \centering
  \includegraphics[ width=0.4\textwidth]{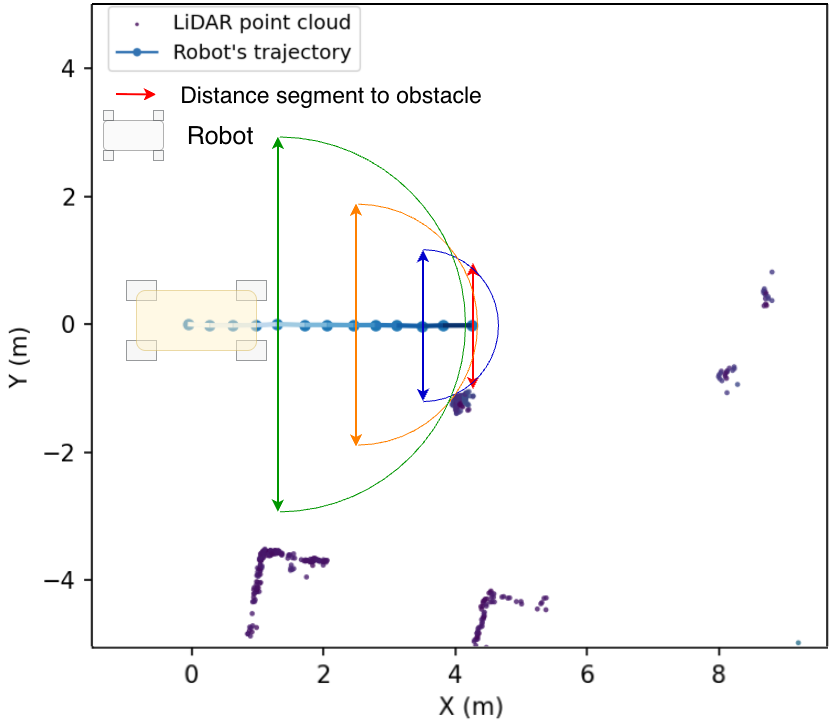}
  \caption{Example clearance computation from Scand \cite{scand}. It is defined for each segment linking two consecutive waypoints as the shortest distance to an obstacle.}
  \label{fig:collision_tunnel}
\end{figure}

We recall that $\mathbf{X}^f$
denotes the fixed-budget LiDAR tokens after the self-attention encoder and $\mathbf{w}=(w,\,\ell)$ the robot width and length. For each candidate positional waypoint $\mathbf{u}_j^{(:2)}\in\mathbb{R}^2$ (with index $j\!=\!1\dots J$), the clearance head applies a single cross-attention layer where trajectory waypoints serves as queries and the LiDAR features act as the keys and values, followed by a small MLP $E_c:\mathbb{R}^d \to \mathbb{R}$ with GELU activation between layers and none at the output applied to each token. $E_c$ outputs a per-waypoint clearance distance:
\begin{align}
\mathbf{q}_j &= W_q\,[\mathbf{u}_j^{(:2)}\,\Vert\,\mathbf{w}],\\
\mathbf{z}_j &= \operatorname{Cr-Att}(Q=\mathbf{q}_j,K=\mathbf{X}^f,V=\mathbf{X}^f)\\
\hat{c}_j &= E_c\big([\mathbf{z}_j\,\Vert\,\mathbf{u}_j^{(:2)}\,\Vert\,\mathbf{w}]\big),
\end{align}
where $d$ is the attention width. 

\paragraph*{Clearance labels generation}
Let $\mathbf{p}_m \in \mathbb{R}^2$, $m=1,\dots,M$, be the LiDAR obstacle points after ROI filtering and ground removal (removing points in a bottom layer of $0.2$ meters).
For each segment
$S_j = [\mathbf{u}_{j-1},\mathbf{u}_j]_{j=1,\dots,J}$, with $\mathbf{u}_0$ the robot's current position,
the point-to-segment distance is:
\begin{equation}
  \begin{aligned}
    \alpha_{m,j} &= \Pi_{[0,1]}\!\left(
      \frac{(\mathbf{p}_m-\mathbf{u}_{j-1})^\top(\mathbf{u}_j-\mathbf{u}_{j-1})}
      {\|\mathbf{u}_j-\mathbf{u}_{j-1}\|^2}
      \right), \\
      &\mathbf{r}_{m,j} =\mathbf{u}_{j-1} + \alpha_{m,j}(\mathbf{u}_j-\mathbf{u}_{j-1}),\\
      &d(\mathbf{p}_m,S_j) = \left\|\mathbf{p}_m-\mathbf{r}_{m,j}\right\|_2.
  \end{aligned}
\end{equation}

where $\Pi_{[0,1]}(x)=\min(1,\max(0,x))$, and $\mathbf{r}_{m,j}$ is the closet point to $\mathbf{p}_m$ on the segment.
The clearance width for each segment is then:
\begin{equation}
d_j = 2 \min_{m=1,\dots,M} d(\mathbf{p}_m,S_j),
\end{equation}
which is represented in Fig. \ref{fig:collision_tunnel}.
From the clearance prediction head, we normalize this distance by the robot size: 
\begin{equation}
    c_j = \frac{d_j}{\max (w, \ell)}
\end{equation}
By keeping this clearance metric greater than one, we ensure that the robot will have a way to avoid the obstacle.

\paragraph*{Loss} The head is trained with an asymmetric Huber Loss:
\begin{equation}
    \begin{aligned}
    e &= c-\hat{c} \\
    \mathrm{Huber}(e) &= \begin{cases}
        \frac{1}{2}(e)^2 &\text{ for } |e|\le \delta\\
        \delta \cdot (|e|-\frac{1}{2}\delta), &\text{ otherwise}
    \end{cases} \\
    w(e) &= \begin{cases}
    w_{\text{over}} & \text{if } e>0,\\
    1 & \text{otherwise,}
    \end{cases} \\
    \mathcal{L}_{\text{clear}} &= w(e) \cdot \mathrm{Huber}(e)
    \end{aligned}
\end{equation} 
which penalizes over-estimation more strongly than under-estimation. This asymmetry enforces a conservative bias: predicting larger clearance (or lower collision risk) than observed is riskier for deployment, so it is weighted more heavily to reduce unsafe optimistic predictions while keeping the robustness of Huber loss to label noise and outliers. We use $\delta=1$ and $w_{\text{over}}=5$.

\subsection{Training objectives and optimization}
\label{subsec:training}
The total objective combines diffusion, clearance prediction, and the auxiliary distance-to-goal regression from the scene token:
\begin{equation}
\mathcal{L}=\lambda_{\text{diff}}\,\mathcal{L}_{\text{diff}}\; +\; \lambda_{\text{clear}}\,\mathcal{L}_{\text{clear}}\; +\; \lambda_{\text{aux}}\,\mathcal{L}_{\text{aux}}.
\end{equation}
We use $\lambda_{\text{diff}}=1\times 10^{-4}$, $\lambda_{\text{clear}}=0.3$ and $\lambda_{\text{aux}}=1-\lambda_{\text{diff}}$.
We use goal-token masking, modality dropout (randomly masking image tokens and predicting clearance only), and robot's size sampling to regularize the shared encoder and prevent degenerate reliance on any single cue. The image and point clouds encoders are frozen during training.

\subsection{Inference and selection}
\label{subsec:inference}

At test time, we sample $N$ waypoint sequences via the reverse diffusion process. For each candidate $n$, the clearance head yields per-step clearance $\{\hat{c}^n_{j}\}_{j=1}^J$. We first define the minimum clearance for each trajectory:
\begin{equation}
    \hat{c}^n = \min_{j\in\{1\dots J\}}\hat{c}^n_{j},
\end{equation}
We identify the set $\mathcal{B}_\text{safe}$ of safe trajectories with a safety margin of twice the robot size:

\begin{equation}
    \mathcal{B}_\text{safe}=\{n: \hat{c}^n>3\}
\end{equation}
If $\mathcal{B}_\text{safe}\neq\varnothing$ we use the trajectory bringing the robot the closest to the goal in $\mathcal{B}_\text{safe}$ with index:
\begin{equation}
    n_g = \arg\min_{n\in\mathcal{B}_\text{safe}}\;||p_g-w^n_J||_2,
\end{equation}
with $p_g$ the goal's position and $w^n_J$ the $J^\text{th}$ waypoint of the trajectory $n$ in the robot's frame.

If $\mathcal{B}_\text{safe}=\varnothing$, we consider the trajectory maximizing the minimum clearance distance with index:
\begin{equation}
    n_c=\arg\max_{n\in\{1\dots N\}}\; \hat{c}^n.
\end{equation}

If $\hat{c}^{n_c} > 1$, we use this trajectory, otherwise we regenerate trajectories by masking the goal to uncondition generation on the 2D objective and focus on the obstacle avoidance task. 
This allows the robot to explore the environment for a safer path without being biased by the goal position when all the ways to the goal seem untraversable. As soon as a feasible way is found with enough confidence, i.e., $\hat{c}^{n_c} > 1.5$ the goal conditioning is unmasked to revert to the goal-reaching behavior.

\section{Evaluation}
\label{sec:experiments}

\subsection{Experimental environment}
\begin{figure}[t]
  \centering
  \includegraphics[width=\linewidth]{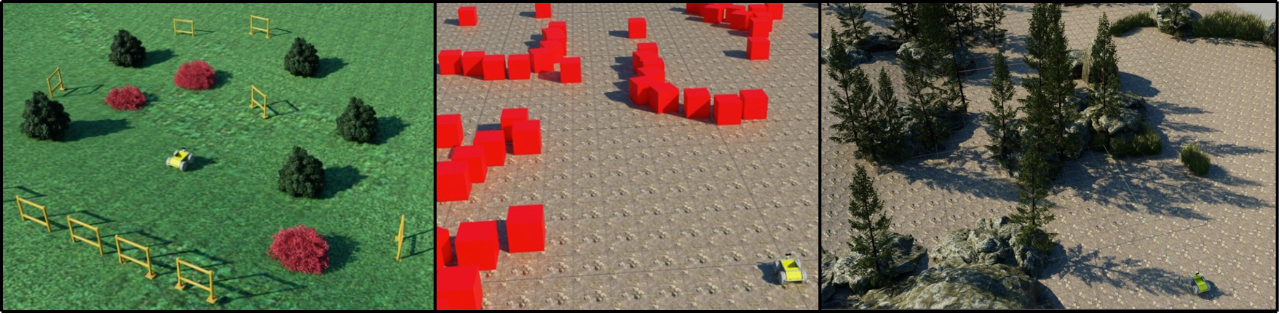}
  \caption{Simulated experimental environments. We sample goals uniformly in the open spaces of the environment within a distance range going from 10 to 50 meters.}
  \label{fig:testing-env}
\end{figure}

\paragraph*{Real-world experiments}
We use a Clearpath Husky equipped with a Zed2i camera and an Ouster OS1 32 channels LiDAR sensor (Fig. \ref{fig:real_tests}, left). The model is executed on an Nvidia AGX Orin embedded computer. The robot's odometry used to compute the relative position of the robot to the goal is computed with Direct Lidar Inertial Odometry~\cite{dlio} using the Ouster's point cloud and IMU signal.
\paragraph*{Simulated experiments}
We use Isaac-Sim \cite{isaac} to create three open-field environments visible in Fig.~\ref{fig:testing-env}, featuring different difficulty levels with bushes, narrow passages, and cluttered obstacles. 
We use the same middle-ware ROS between the sensor signals provided by the simulation and our model to better match real-world deployment conditions.
We use a Clearpath Husky rover with a front RGB camera and a 32-beam Ouster 3D LiDAR similarly to our real-world platform. For each environment, we sample 100 goals uniformly over free spaces to obtain the metrics in Table~\ref{tab:main-results}. Episodes terminate on \textit{success} (goal within 0.5 m radius), 
\textit{collision}, or a time exceeding 10 minutes without reaching the goal's position.
\paragraph*{Low level control}
In both configurations the policy output's second waypoint is tracked by a low-level Proportional velocity controller. The position waypoint as well as the angular reference are converted to linear and rotational speed commands that are sent to the robot or the simulation interface in order to be converted to wheel rotation speed commands.

\subsection{Training Data}
\label{sec:train-data}
\begin{figure*}[t]
  \centering
  \includegraphics[width=.95\textwidth]{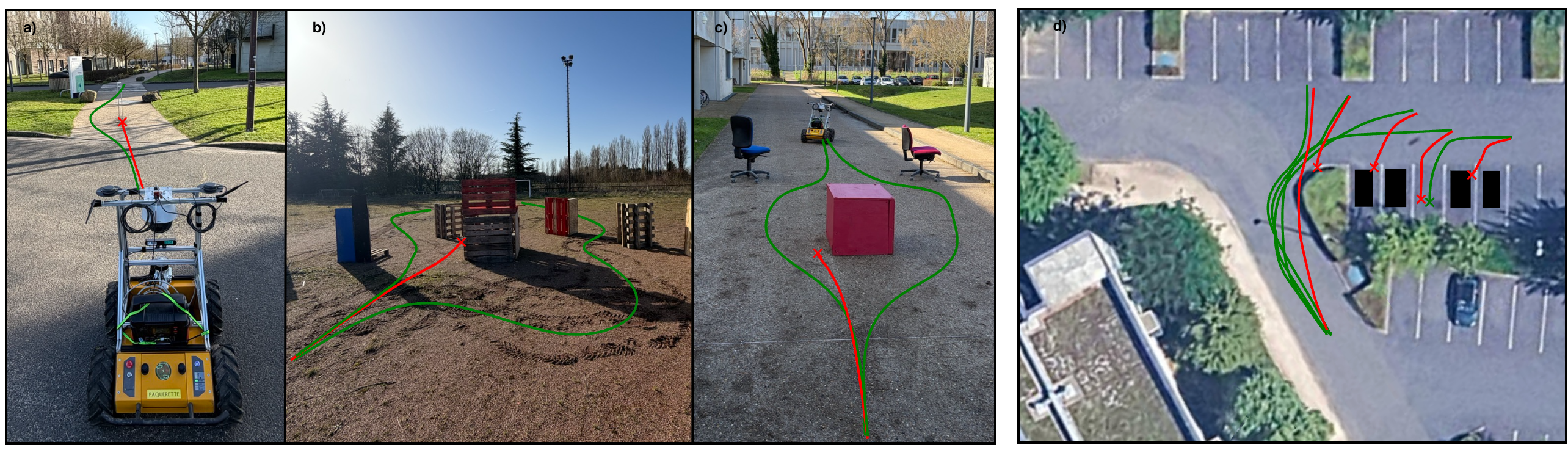}
  \caption{Zero-shot deployment of ViLiNT (green curves) and NoMaD-FT (Red curves). We compare the capability of both solutions to prevent collision occurrences while navigating our robot towards 2D goals in environments and obstacles unknown from our training data. Goals are set to be 30 m away from the initial robot's position.}
  \label{fig:real_tests}
\end{figure*}

We aggregate a heterogeneous corpus of publicly available and custom datasets (Table~\ref{tab:train-data}). We add data from our simulated environment (Env.1 only in Fig.~\ref{fig:testing-env}) in order to reach reasonable performance as the distribution shift between our evaluation environment and the other datasets is high. We also add data recorded from our robotic platform, in environment without the test obstacles (Fig.~\ref{fig:real_tests}) so that the model can learn from features corresponding to our sensor type. To match state-of-the-art standards in terms of data quantity for training \cite{gnm,vint,nomad}, we use datasets lacking LiDAR sensor data such as Huron. To integrate such datasets in the training process, we mask the LiDAR tokens and do not compute clearance predictions.
Images, LiDAR point clouds, and poses from these trajectories are sampled at a period $\Delta t{=}0.25\,\mathrm{s}$. 
To ensure consistent learning of trajectory diffusion across datasets, we adapt the temporal spacing between waypoints on a per-dataset basis. 
For a dataset $d$ with average spatial waypoint spacing $\delta_{\mathrm{wp}}^d$, we define an integer rescaling factor
\begin{equation}
k_d=\left\lfloor \frac{\underset{d}{\max}\{\delta_{\mathrm{wp}}^d\}}{\delta_{\mathrm{wp}}^d} \right\rfloor.
\end{equation}
The effective waypoint sampling period used for dataset $d$ is then $k_d \Delta t$, with $\Delta t=0.25\,\mathrm{s}$. This discretized rescaling makes the average metric distance between consecutive waypoints more comparable across datasets, yielding a more homogeneous trajectory distribution for diffusion training.

\begin{table}[t]
\caption{Training data summary.}
\label{tab:train-data}
\centering
\setlength{\tabcolsep}{3.5pt}
\begin{tabular}{lccc}
\hline
\textbf{Dataset} & \textbf{Modalities} & \textbf{Hours} & \textbf{Terrain} \\
\hline
RELLIS-3D \cite{rellis3d}             & RGB+LiDAR & 35 min & Off-Road  \\
TartanDrive 2.0 \cite{tartandrive2}   & RGB+LiDAR & 7h & Off-Road  \\
SCAND \cite{scand}                    & RGB+LiDAR & 9h & Sidewalks  \\
Husky Off-Road (ours)                 & RGB+LiDAR & 30 min & Sidewalks  \\
Isaac-Sim (ours)                      & RGB+LiDAR & 30 min & Off-Road simulation  \\
Huron                                 & RGB+2D scans & 75h & Sidewalks \\
Grand-tour \cite{grand-tour}          & RGB+LiDAR & 6h & Sidewalks + Off-road \\
\hline
\textbf{Total}                        &           & 98h35 &  \\
\hline
\end{tabular}
\vspace{-0.5em}
\end{table}

\subsection{Baselines and metrics}
We evaluate our model (\modelname) against other end-to-end and classical baselines chosen to be representative of different approach families:
\begin{itemize}
  \item \textbf{NoMaD-FT} We fine-tune the network on our data, switching the goal-image encoder to a 2D goal-position encoder following \cite{nomad} that states that goal encoding can be switched without degrading navigation performances. We fine-tune the model for 20 epochs on our dataset using the hyperparameters recommended by \cite{nomad}.
  \item \textbf{NoMaD-Col} We retrain NoMaD using the image output tokens to predict the clearance distance and compute the collision heuristic at inference. This version allows us to evaluate the feasibility of predicting clearance distances from image tokens only.
  \item \textbf{TEB-Elev} As a non-learning baseline, we use the Timed-Elastic-Band local planner \cite{teb} applied to a local elevation map \cite{elevation} built from the LiDAR data and thresholded to detect obstacles. The global planning is done using an A$^*$ algorithm.
  \item  \textbf{\modelname{}-Image-Mask} and \textbf{\modelname{}-LiDAR-Mask} are our models when we mask images and LiDAR, respectively, during inference.
\end{itemize}

We use the following metrics, with $S_i$ the success of the trial (binary), $c_i$ the binary collision indicator at trial $i$:
\begin{itemize}
    \item SR: Success Rate over a sufficient number of trials allows us to correctly evaluate our policy's performance: \\ $\text{SR}=\frac{1}{N}\sum_{i=1}^{N} S_i$
    \item CR: Collision Rate measures the average number of collisions during episodes: \\ $\text{CR}=\frac{1}{N}\sum_{i=1}^{N} c_i$
\end{itemize}

\subsection{Results}
\label{sec:main-results}

\paragraph{Real-world}
First, we validate qualitatively the  capability of our model to track long-range goals. To do so, we set goals more than 100 meters away from the initial position of our robot and ensured that \textbf{\modelname{}} was able to reach that goal avoiding obstacles of reasonable size in between (size comparable to obstacles we can see in Fig.~\ref{fig:real_tests}).
Then, to show more specifically the increased obstacle avoidance capabilities of \textbf{\modelname{}} we set goals 30 m behind  obstacle fields and compare the behaviors of our solution and \textbf{NoMaD-FT} (Fig.~\ref{fig:real_tests}). \textbf{\modelname{}} features a 85\% \textbf{SR} whereas \textbf{NoMaD-FT} scores a 15\% \textbf{SR}. \textbf{\modelname{}} consistently outperforms image-only models at avoiding obstacles. When the image model manages not to collide with an obstacle, we observe that the robot's path comes very close to the obstacle and that the model does not take advantage of the free space to track a safer trajectory whereas \textbf{\modelname{}} does. In the parking environment (Fig.~\ref{fig:real_tests}, right), we observe the \modelname{} consistently circumvent obstacles, except when it enters an empty car spot, showing the limit of local planning.

\begin{table}[ht]
\caption{Combined navigation and safety metrics across simulated off-road environments. Mean over 100 runs; higher is better for SR, lower is better for CR.}
\label{tab:main-results}
\centering
\resizebox{\linewidth}{!}{%
\begin{tabular}{l cc|cc|cc}
\hline
\textbf{Method}
& \multicolumn{2}{c|}{\textbf{Env. 1}}
& \multicolumn{2}{c|}{\textbf{Env. 2}}
& \multicolumn{2}{c}{\textbf{Env. 3}} \\
\cline{2-7}
& \textbf{SR}~$\uparrow$ &  \textbf{CR}~$\downarrow$
& \textbf{SR}~$\uparrow$ &  \textbf{CR}~$\downarrow$
& \textbf{SR}~$\uparrow$ &  \textbf{CR}~$\downarrow$ \\
\hline
NoMaD-FT\cite{nomad}
& 0.33 & 0.67
& 0.37 & 0.63
& 0.12  & 0.88 \\
NoMaD-Col 
& 0.20 & 0.80
& 0.38 & 0.62
& 0.19  & 0.81 \\
\modelname{}-LiDAR-Mask
& 0.17 & 0.83
& 0.37 & 0.63
& 0.20  & 0.80 \\
\modelname{}-Image-Mask
& 0.72 & 0.28
& \textbf{0.67} & \textbf{0.33}
& 0.23  & 0.77 \\
\textbf{\modelname{} (ours)}
& \textbf{0.79} & \textbf{0.21}
& 0.61 & 0.39
& \textbf{0.76} & \textbf{0.24}\\
\hline
TEB-Elev
& 0.66  & 0.00
& 0.89  & 0.00
& 1.00 & 0.00 \\
\hline
\end{tabular}}
\vspace{-0.6em}
\end{table}

\paragraph{Simulation}
Table~\ref{tab:main-results} summarizes our model's performance in comparison to our baselines. \textbf{\modelname} significantly improves navigation performance relative to unimodal policies over every test environment with a 166\% increase in \textbf{SR} and a 62\% reduction in \textbf{CR} compared to \textbf{NoMaD-FT}. 

We can also see in Table~\ref{tab:main-results} that image policies (NoMaD-FT, NoMaD-Col, ViLiNT-LiDAR-Mask) have similar results within 1\% accuracy. Bringing LiDAR data (\modelname{}-Image-Mask) in the architecture grants a 100\% \textbf{SR} increase and a 37\% reduction in \textbf{CR}. 
We additionally gain 33\% in \textbf{SR} and  decrease 39\% in \textbf{CR} when using the image sequence besides LiDAR in input (\modelname{} vs \modelname{}-Image-Mask).

We note that the TEB-Elev baseline, which is configured to plan optimal paths toward the goal by maintaining a safety margin from obstacles, outperforms all learning-based methods. No collisions are observed when running TEB-Elev, and on average, the SR achieves 85\% over all three environments. Some runs still fail due to the conservative safety margin causing the planner to get stuck or fail to find feasible paths, in particular in Env. 1, where obstacles are closer to each other. This could be improved, but this shows that despite its superior performance, TEB indeed requires extensive parameter tuning to adapt to different robots or environments.

\paragraph{Embodiment influence}

\begin{figure}[t]
  \centering
  \includegraphics[width=0.45\textwidth]{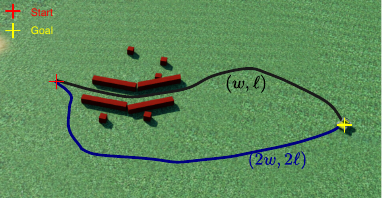}
  \caption{Simulated experience showing how changing the robot's size input influences the navigation behavior.}
  \label{fig:size-test}
\end{figure}
In the simple scenario illustrated in Figure~\ref{fig:size-test}, we compare deployment with the nominal robot dimensions encoded in the embodiment vector 
and deployment with dimensions scaled by a factor of two. The goal is placed 20m ahead along the robot’s initial $x$-axis. For the nominal size, the 
selected trajectory passes through the corridor formed by the red blocks, whose width is $1.5\max(w,\ell)$.
When the robot dimensions are inflated, trajectories crossing this corridor are rejected by the clearance heuristic, and the robot instead bypasses 
the obstacle field on the right. Note that this avoidance behavior is limited by the variety of generated trajectories that always move forward, meaning 
that if we continue increasing the robot size, the robot will eventually get stuck because no safe trajectory will be generated.

\section{Discussion and Conclusion}
\label{sec:discussion}

We introduced a multimodal navigation transformer that fuses RGB images, 3D LiDAR data, and platform description tokens to enable robust zero-shot 
navigation across ground robots of varying sizes. By combining token-level multimodal fusion, diffusion-based trajectory generation, and a 
robot-size-conditioned clearance prediction head, the model achieves both expressive planning (via diffusion sampling) and selective decision-making 
(via risk-based ranking). Experimental results in real and simulated off-road environments show improved goal-reaching efficiency and reduced collisions, 
despite limited training data.

However, trajectory selection remains decoupled from generation and relies on thresholded goal masking to escape local minima, 
which can be conservative and induce exploratory detours. Future work will focus on joint optimization of trajectory generation and 
clearance prediction (e.g., reinforcement learning with collision feedback), improved traversability and clearance estimation, and 
robot-specific local path planning strategies.

\bibliographystyle{IEEEtran} 
\bibliography{references}

@article{roadrunner,
  author    = {Frey, J. and Patel, M. and Atha, D. and Nubert, J. and Fan, D. and Agha, A. and others},
  title     = {Roadrunner-Learning Traversability Estimation for Autonomous Off-Road Driving},
  journal   = {IEEE Transactions on Field Robotics},
  volume    = {1},
  pages     = {192--212},
  year      = {2024}
}

@inproceedings{terrainnet,
  author    = {Meng, X. and Hatch, N. and Lambert, A. and Li, A. and Wagener, N. and others},
  title     = {{TerrainNet}: Visual Modeling of Complex Terrain for High-speed, Off-road Navigation},
  booktitle = {Robotics: Science and Systems},
  year      = {2023}
}

@inproceedings{vstrong,
  author    = {Jung, S. and Lee, J. and Meng, X. and Boots, B. and Lambert, A.},
  title     = {{V-STRONG}: Visual Self-Supervised Traversability Learning for Off-Road Navigation},
  booktitle = {2024 IEEE International Conference on Robotics and Automation (ICRA)},
  year      = {2024},
  month     = may
}

@article{mppi,
  author    = {Williams, G. and Aldrich, A. and Theodorou, E. A.},
  title     = {Model Predictive Path Integral Control: From Theory to Parallel Computation},
  journal   = {Journal of Guidance, Control, and Dynamics},
  volume    = {40},
  number    = {2},
  pages     = {344--357},
  year      = {2017}
}

@inproceedings{teb,
  author    = {R{\"o}smann, C. and Hoffmann, F. and Bertram, T.},
  title     = {Timed-Elastic-Bands for Time-Optimal Point-to-Point Nonlinear Model Predictive Control},
  booktitle = {2015 European Control Conference (ECC)},

  year      = {2015},
  doi       = {10.1109/ECC.2015.7331052}
}

@inproceedings{traversability-costmap2,
  author    = {Castro, M. G. and Triest, S. and Wang, W. and Gregory, J. M. and Sanchez, F. and others},
  title     = {How Does It Feel? Self-Supervised Costmap Learning for Off-Road Vehicle Traversability},
  booktitle = {IEEE International Conference on Robotics and Automation (ICRA)},
  year      = {2023}
}

@inproceedings{gnm,
  author    = {Shah, D. and Sridhar, A. and Bhorkar, A. and Hirose, N. and Levine, S.},
  title     = {{GNM}: A General Navigation Model to Drive Any Robot},
  booktitle = {2023 IEEE International Conference on Robotics and Automation (ICRA)},
  address   = {London, United Kingdom},
  pages     = {7226--7233},
  year      = {2023}
}

@inproceedings{vint,
  author    = {Shah, D. and Sridhar, A. and Dashora, N. and Stachowicz, K. and Black, K. and others},
  title     = {{ViNT}: A Foundation Model for Visual Navigation},
  booktitle = {Conference on Robot Learning},
  publisher = {PMLR},
  year      = {2023}
}

@inproceedings{nomad,
  author    = {Sridhar, A. and Shah, D. and Glossop, C. and Levine, S.},
  title     = {{NoMaD}: Goal Masked Diffusion Policies for Navigation and Exploration},
  booktitle = {2024 IEEE International Conference on Robotics and Automation (ICRA)},
  pages     = {63--70},
  publisher = {IEEE},
  year      = {2024},
  month     = may
}

@inproceedings{dtg,
  author    = {Liang, J. and Payandeh, A. and Song, D. and Xiao, X. and Manocha, D.},
  title     = {{DTG}: Diffusion-Based Trajectory Generation for Mapless Global Navigation},
  booktitle = {2024 IEEE/RSJ International Conference on Intelligent Robots and Systems (IROS)},
  pages     = {5340--5347},
  publisher = {IEEE},
  year      = {2024},
  month     = oct
}

@inproceedings{agile,
  author    = {Pan, Y. and Cheng, C. A. and Saigol, K. and Lee, K. and Yan, X. and Theodorou, E. and Boots, B.},
  title     = {Agile Autonomous Driving using End-to-End Deep Imitation Learning},
  booktitle = {Robotics: Science and Systems},
  year      = {2017}
}

@inproceedings{rellis3d,
  author    = {Jiang, P. and Osteen, P. and Wigness, M. and Saripalli, S.},
  title     = {{RELLIS-3D} Dataset: Data, Benchmarks and Analysis},
  booktitle = {2021 IEEE International Conference on Robotics and Automation (ICRA)},

  year      = {2021},
  doi       = {10.1109/ICRA48506.2021.9561251}
}

@inproceedings{tartandrive2,
  author    = {Sivaprakasam, M. and Maheshwari, P. and Castro, M. G. and Triest, S. and Nye, M. and others},
  title     = {{TartanDrive} 2.0: More Modalities and Better Infrastructure to Further Self-Supervised Learning Research in Off-Road Driving Tasks},
  booktitle = {IEEE International Conference on Robotics and Automation (ICRA)},
  year      = {2024},
  month     = may
}

@inproceedings{risk_guided_diffusion,
  author    = {Thakker, R. and Patnaik, A. and Kurtz, V. and Frey, J. and Becktor, J. and others},
  title     = {Risk-Guided Diffusion: Toward Deploying Robot Foundation Models In Space, Where Failure Is Not An Option},
  booktitle = {RSS Workshop on Reliable Robotics: Safety and Security in the Face of Generative AI},
  year      = {2025}
}

@article{mono_mpc,
  title={MonoMPC: Monocular Vision Based Navigation With Learned Collision Model and Risk-Aware Model Predictive Control},
  author={Sharma, Basant and Jadhav, Prajyot and Paul, Pranjal and Krishna, K Madhava and Singh, Arun Kumar},
  journal={IEEE Robotics and Automation Letters},
  volume={11},
  number={2},
  pages={1330--1337},
  year={2025},
  publisher={IEEE}
}

@inproceedings{pact,
  author    = {Bonatti, R. and Vemprala, S. and Ma, S. and Frujeri, F. and Chen, S. and Kapoor, A.},
  title     = {{PACT}: Perception-Action Causal Transformer for Autoregressive Robotics Pre-Training},
  booktitle = {2023 IEEE/RSJ International Conference on Intelligent Robots and Systems (IROS)},
  pages     = {3621--3627},
  year      = {2022}
}

@inproceedings{pointnet,
  author    = {Qi, C. R. and Su, H. and Mo, K. and Guibas, L. J.},
  title     = {{PointNet}: Deep Learning on Point Sets for 3D Classification and Segmentation},
  booktitle = {IEEE Conference on Computer Vision and Pattern Recognition (CVPR)},
  year      = {2017}
}

@inproceedings{minkowskiengine,
  author    = {Choy, C. and Gwak, J. and Savarese, S.},
  title     = {4D Spatio-Temporal ConvNets: Minkowski Convolutional Neural Networks},
  booktitle = {IEEE/CVF Conference on Computer Vision and Pattern Recognition (CVPR)},
  pages     = {3075--3084},
  year      = {2019}
}

@article{visuomotordiffusion,
  author  = {Chi, C. and Xu, Z. and Feng, S. and Cousineau, E. and Du, Y. and others},
  title   = {Diffusion Policy: Visuomotor Policy Learning via Action Diffusion},
  journal = {The International Journal of Robotics Research},
  year    = {2023}
}

@misc{rl_diffusion_planning,
  author        = {Lee, G. and Park, D. and Jeong, J. and Yoon, K. J.},
  title         = {Non-differentiable Reward Optimization for Diffusion-based Autonomous Motion Planning},
  year          = {2025},
  eprint        = {2507.12977},
  archivePrefix = {arXiv},
  url           = {https://arxiv.org/abs/2507.12977},
}

@inproceedings{polarnet,
  author    = {Zhang, Y. and Zhou, Z. and David, P. and Yue, X. and Xi, Z. and others},
  title     = {{PolarNet}: An Improved Grid Representation for Online LiDAR Point Clouds Semantic Segmentation},
  booktitle = {IEEE/CVF Conference on Computer Vision and Pattern Recognition (CVPR)},
  pages     = {9601--9610},
  year      = {2020}
}

@article{scand,
  author  = {Karnan, H. and Nair, A. and Xiao, X. and Warnell, G. and Pirk, S. and others},
  title   = {Socially Compliant Navigation Dataset ({SCAND}): A Large-Scale Dataset of Demonstrations for Social Navigation},
  journal = {IEEE Robotics and Automation Letters},
  volume  = {7},
  number  = {4},
  pages   = {11807--11814},
  year    = {2022}
}

@misc{isaac,
author = {{NVIDIA}},
license = {Apache-2.0},
title = {{Isaac Sim}},
url = {https://github.com/isaac-sim/IsaacSim},
version = {5.1.0}
}

@misc{dave2,
  author        = {Bojarski, M. and Del Testa, D. and Dworakowski, D. and Firner, B. and Flepp, B. and others},
  title         = {End to End Learning for Self-Driving Cars},
  year          = {2016},
  eprint        = {1604.07316},
  archivePrefix = {arXiv},
  url           = {https://arxiv.org/abs/1604.07316},
}

@inproceedings{elevation,
  author    = {Miki, T. and Wellhausen, L. and Grandia, R. and Jenelten, F. and Homberger, T. and Hutter, M.},
  title     = {Elevation Mapping for Locomotion and Navigation using {GPU}},
  booktitle = {2022 IEEE/RSJ International Conference on Intelligent Robots and Systems (IROS)},
  publisher = {IEEE},
  year      = {2022}
}

@inproceedings{dune,
  author    = {Sar{\i}y{\i}ld{\i}z, Mert B{\"u}lent and Weinzaepfel, Philippe and Lucas, Thomas and De Jorge, Pau and Larlus, Diane and Kalantidis, Yannis},
  title     = {{DUNE}: Distilling a Universal Encoder from Heterogeneous 2D and 3D Teachers},
  booktitle = {IEEE/CVF Conference on Computer Vision and Pattern Recognition  (CVPR)},
  pages     = {30084--30094},
  year      = {2025}
}

@inproceedings{pointtransformerv3,
  author    = {Wu, X. and Jiang, L. and Wang, P. S. and Liu, Z. and Liu, X. and Qiao, Y. and Ouyang, W. and He, T. and Zhao, H.},
  title     = {Point Transformer {V3}: Simpler, Faster, Stronger},
  booktitle = {Advances in Neural Information Processing Systems},
  year      = {2023}
}

@inproceedings{dlio,
  author    = {Chen, K. and Nemiroff, R. and Lopez, B. T.},
  title     = {Direct {LiDAR-Inertial} Odometry: Lightweight {LIO} with Continuous-Time Motion Correction},
  booktitle = {2023 IEEE International Conference on Robotics and Automation (ICRA)},
  address   = {London, United Kingdom},
  pages     = {3983--3989},
  year      = {2023},
  doi       = {10.1109/ICRA48891.2023.10160508}
}

@misc{grand-tour,
  title         = {GrandTour: A Legged Robotics Dataset in the Wild for Multi-Modal Perception and State Estimation},
  author        = {Jonas Frey and Turcan Tuna and Frank Fu and Katharine Patterson and Tianao Xu and Maurice Fallon and Cesar Cadena and Marco Hutter},
  year          = {2026},
  eprint        = {2602.18164},
  archivePrefix = {arXiv},
  primaryClass  = {cs.RO},
  url           = {https://arxiv.org/abs/2602.18164},
}

\end{document}